\documentclass[accepted]{uai2021} 

\usepackage[american]{babel}

\usepackage{natbib} 
\usepackage{mathtools} 
\usepackage{booktabs} 
\usepackage{tikz} 
\usepackage{amsmath,amssymb}
\usepackage{bbold}
\usepackage{bm}
\usepackage[linesnumbered,ruled,vlined]{algorithm2e}
\usepackage{todonotes}
\usepackage{enumitem}

\usepackage{xr}
\graphicspath{{./figures/}}

\title{\LARGE High-Dimensional Bayesian Optimization with Sparse Axis-Aligned Subspaces}

\author[1]{\href{mailto:David Eriksson <deriksson@fb.com>?Subject=Your UAI 2021 paper}{David Eriksson\thanks{Equal contribution}\textsuperscript{,}}{}} 
\author[2]{Martin Jankowiak$^{*,}$}
\affil[1]{%
    Facebook\\
    Menlo Park, California, USA
}
\affil[2]{%
    Broad Institute of Harvard and MIT\\
    Cambridge, Massachusetts, USA
}

\newcommand{\fobj}{f_{\text{obj}}}
\newcommand{\algoname}{{\texttt {SAASBO}}}
\newcommand{\priorname}{{\texttt {SAAS}}}
\newcommand{\by}{\mathbf{y}}

\newcommand{\bff}{\mathbf{f}}
\newcommand{\bx}{\mathbf{x}}
\newcommand{\bz}{\mathbf{z}}
\newcommand{\bX}{\mathbf{X}}
\newcommand{\EE}{\mathbb{E}}
\newcommand{\DD}{\mathcal{D}}
\newcommand{\OO}{\mathcal{O}}
\newcommand{\NN}{\mathcal{N}}

\begin{document}
\maketitle

\begin{abstract}
    Bayesian optimization (BO) is a powerful paradigm for efficient optimization of black-box objective functions. High-dimensional BO presents
    a particular challenge, in part because the curse of dimensionality makes it difficult to define---as well as do inference over---a suitable class of surrogate models.
    We argue that Gaussian process surrogate models defined on sparse axis-aligned subspaces offer an attractive compromise between flexibility and parsimony.
    We demonstrate that our approach, which relies on Hamiltonian Monte Carlo for inference, can rapidly identify sparse subspaces relevant to modeling the unknown objective function, enabling sample-efficient high-dimensional BO.
    In an extensive suite of experiments comparing to existing methods for high-dimensional BO we demonstrate that our algorithm, Sparse Axis-Aligned Subspace BO (\algoname{}), achieves excellent performance on several synthetic and real-world problems without the need to set problem-specific hyperparameters.
\end{abstract}

\section{Introduction}
\label{sec:intro}

Optimization plays an essential role in many fields of science, engineering and beyond.
From calibrating complex experimental systems to tuning hyperparameters
of machine learning models,
the need for scalable and efficient optimization methods is ubiquitous.
Bayesian Optimization (BO) algorithms have proven particularly successful on a wide variety of domains including hyperparameter tuning~\citep{snoek2012practical}, A/B tests~\citep{letham2019constrained}, chemical engineering~\citep{hernandez2017parallel}, materials science~\citep{ueno2016combo},
control systems~\citep{candelieri2018bayesian},
and drug discovery~\citep{negoescu2011knowledge}.

These algorithms typically consist of two components.
The first component employs Bayesian methods to construct a surrogate model of the (unknown) objective function.
The second component uses this model together with an acquisition function to select the most promising query point(s) at which to evaluate the objective function.
By leveraging the uncertainty quantification provided by the Bayesian model, a well-designed BO algorithm can provide an effective balance between exploration and exploitation, leading to highly sample-efficient optimization.

While BO has become a workhorse algorithm that is employed in a wide variety of settings, successful applications are often limited to low-dimensional problems, e.g.~fewer than twenty dimensions~\citep{frazier2018tutorial}.
Applying BO to high-dimensional problems remains a significant challenge.
The difficulty can be traced to both of the algorithm components mentioned above,
although we postulate that suitable function priors are especially important for
good performance.
In particular, in order for BO to be sample-efficient in high-dimensional spaces, it is crucial to define surrogate models that are sufficiently parsimonious that they can be inferred from a small number of query points.
An overly flexible class of models is likely to suffer from overfitting, which severely limits its effectiveness in decision-making.
Likewise, an overly rigid class of models is unlikely to capture enough features of the objective function.
A compromise between flexibility and parsimony is essential.

In this work we focus on the setting where we aim to optimize a black-box function with hundreds of variables and where we are limited to a few hundred queries of the objective function.
We argue that in this low-sample regime Gaussian process surrogate models defined on sparse axis-aligned subspaces provide an attractive compromise between flexibility and parsimony.
More specifically, our contributions are as follows:
\begin{itemize}
    \item We propose the sparsity-inducing \priorname{} function prior
    \item We demonstrate that when combined with the No-Turn-U-Sampler (NUTS) for inference,
        our surrogate model quickly identifies the most relevant low-dimensional subspace, which in turn leads to sample-efficient BO.
    \item We show that \algoname{}
        outperforms a number of strong baselines on several problems, including three real-world problems with as many as 388 dimensions, all without setting problem-specific hyperparameters.
\end{itemize}

\section{Related Work}
\label{sec:related}
There is a large body of research on high-dimensional BO, and a wide variety
of surrogate modelling and acquisition strategies have been proposed \citep{chen2012joint}. In
the following we draw attention to a number of common themes.

A popular approach is to rely on low-dimensional structure, with several methods utilizing random projections~\citep{wang2016bayesian,qian2016derivative, binois2020choice,letham2020re}.
REMBO uses a random projection to project low-dimensional points up to the original space~\citep{wang2016bayesian}.
ALEBO introduces several refinements to REMBO and demonstrates improved performance across a large number of problems~\citep{letham2020re}.
Alternatively, the embedding can be learned jointly with the model, including
both linear ~\citep{garnett2013active} and non-linear~\citep{lu2018structured} embeddings.
Finally, Hashing-enhanced Subspace BO (HeSBO)~\citep{nayebi2019framework} relies on hashing and sketching to reduce surrogate modeling and acquisition function optimization to a low-dimensional space.

Several methods rely on additive structure, where the function is assumed to be a sum of low-dimensional components~\citep{kandasamy2015high, gardner2017discovering,mutny2019efficient,wang2018batched}.
This approach allows separating the input space into independent domains, reducing the effective dimensionality of the model.

A common feature of many BO algorithms in high dimensions is that they tend to prefer highly uncertain query points near the domain boundary.
As this is usually where the model is the most uncertain, this is often a poor choice that leads to over-exploration and poor optimization performance.
\citet{oh2018bock} address this issue by introducing a cylindrical kernel that promotes
selection of query points in the interior of the domain.
LineBO~\citep{kirschner2019adaptive} optimizes the acquisition function along one-dimensional lines, which also helps to avoid highly uncertain points.
The TuRBO algorithm uses several trust-regions centered around the current best solution \citep{turbo}.
These trust-regions are resized based on progress, allowing TuRBO to zoom-in on promising regions.
\citet{li2018high} use dropout to select a subset of dimensions over which to optimize the acquisition function, with excluded dimensions fixed to the value of the best point found so far.

Most similar to our method is COMBO \citep{oh2019combinatorial}, which uses a sparsity-inducing prior in conjunction with a finite feature expansion to define a surrogate model that is suitable for BO on combinatorial search spaces. The finite feature expansion enables efficient inference via slice sampling. Unfortunately, a finite feature expansion is inappropriate in our setting with real-valued inputs, since the curse of dimensionality severely limits the flexibility of the resulting function prior.

It also important to note that there are many black-box optimization algorithms that
do not rely on Bayesian methods, with evolutionary algorithms being especially common.
While most methods require thousands of evaluations to find good minima~\citep{yu2010introduction},
the popular covariance matrix adaptation evolution strategy (CMA-ES; ~\citep{hansen2003reducing}) is competitive with BO on some problems~\citep{letham2020re}.

\section{Background}
\label{sec:bg}

We use this section to establish our notation and review necessary background material.
Throughout this paper we work in the $D$-dimensional domain $\DD = [0, 1]^D$.
We consider the minimization problem $\bx_{\rm min} \in {\rm argmin}_{\bx \in \DD} \,\fobj(\bx)$ for a noise-free
objective function $\fobj: \DD \to \mathbb{R}$.
We assume that evaluations of $\fobj$ are costly and that we are limited to at most a few hundred.
Additionally, $\fobj$ is a black-box function and gradient information is unavailable.

The rest of this section is organized as follows:
in Sec.~\ref{sec:gp} we review Gaussian processes; and in Sec.~\ref{sec:ei} we review the expected improvement
acquisition function.

\subsection{Gaussian Processes}
\label{sec:gp}

Gaussian processes (GPs) offer powerful non-parametric function priors that are the gold standard
in BO due to their flexibility and excellent uncertainty quantification.
A GP on the input space $\DD$
is specified\footnote{Here and elsewhere we assume that the mean function is uniformly zero.}
 by a covariance function or kernel $k:\DD\times \DD \to \mathbb{R}$ \citep{rasmussen2003gaussian}.
 A common choice is the RBF or squared exponential kernel, which is given by
\begin{equation}
k^\psi(\bx, \by) = \sigma_k^2 \exp \{ -\tfrac{1}{2} \sum_i \rho_i (x_i - y_i)^2 \}
\end{equation}
where $\rho_i$ for $i=1,...,D$ are inverse squared length scales
and where we use $\psi$ to collectively denote all the hyperparameters, i.e.~$\psi = \{\rho_{1:D}, \sigma_k^2\}$.
For scalar regression $f: \DD \to \mathbb{R}$ the joint density of a GP takes the form
\begin{equation}
p(\by, \bff | \bX) = \NN(\by|\bff, \sigma^2 \mathbb{1}_N) \NN(\bff | \bm{0}, K^{\psi}_{\bX\bX})
\end{equation}
where $\by$ are the real-valued targets, $\bff$ are the latent function values,
$\bX = \{ \bx_i \}_{i=1}^N$ are the $N$ inputs with $\bx_i \in \DD$,
$\sigma^2$ is the variance of the Normal likelihood $\NN(\by|\cdot)$,
and $K^{\psi}_{\bX\bX}$ is the $N \times N$ kernel matrix.
Throughout this paper we will be interested in modeling noise-free functions,
in which case $\sigma^2$ is set to a small constant.
The marginal likelihood of the observed data can be computed in closed form:
\begin{align}
\label{eqn:mll}
p(\by|\bX, \psi) = \int \! d \bff \; p(\by, \bff | \bX)    
                = \NN(\by, K^{\psi}_{\bX\bX} + \sigma^2 \mathbb{1}_N).
\end{align}
The posterior distribution of the GP at a query point $\bx^* \in \DD$ is the
Normal distribution $\NN(\mu_\bff(\bx^*), \sigma_\bff(\bx^*)^2)$
where $\mu_\bff(\cdot)$ and $\sigma_\bff(\cdot)^2$ are given by
\begin{align}
\label{eqn:gppredmean}
\mu_\bff(\bx^*) &= {k^{\psi}_{* \bX}}^{\rm T}  {(K^{\psi}_{\bX\bX} + \sigma^2 \mathbb{1}_N)}^{-1}\by    \\
\label{eqn:gppredvar}
\sigma_\bff(\bx^*)^2 &=  k^{\psi}_{**} - {k^{\psi}_{* \bX}}^{\rm T}  {(K^{\psi}_{\bX\bX} + \sigma^2 \mathbb{1}_N)}^{-1}    {k^{\psi}_{* \bX}}
\end{align}
Here $k^{\psi}_{**} =  k^\psi(\bx^*, \bx^*) $ and ${k^{\psi}_{* \bX}}$ is the column vector specified by $({k^{\psi}_{* \bX}})_n=k^\psi(\bx^*, \bx_n)$ for
$n=1,...,N$.

\subsection{Expected Improvement}
\label{sec:ei}

Expected improvement (EI) is a popular acquisition function that is defined as follows \citep{mockus1978toward, jones1998efficient}.
Suppose that in previous rounds of BO we have collected $\mathcal{H} = \{\bx_{1:N}, y_{1:N}\}$.
Then let $y_{\rm min} = {\rm min}_n \, y_n$ denote the best function
evaluation we have seen so far.
We define the \emph{improvement} $u(\bx | y_{\rm min})$ at query point $\bx \in \DD$ as
$u(\bx | y_{\rm min}) = {\rm max}(0, y_{\rm min} - f(\bx))$.
EI is defined as the expectation of the improvement over the posterior of $f(\bx)$:
\begin{align}
\label{eqn:eidef}
\rm{EI}(\bx |  y_{\rm min}, \psi) = \EE_{p(f(\bx) | \psi, \mathcal{H})} \left[ u(\bx | y_{\rm min}) \right]
\end{align}
 where our notation makes explicit the dependence of Eqn.~\eqref{eqn:eidef} on the kernel hyperparameters $\psi$.
For a GP like in Sec.~\ref{sec:gp} this expectation can be evaluated in closed form:
\begin{align}
\label{eqn:eiclosedform}
\rm{EI}(\bx |  y_{\rm min}, \psi) = (y_{\rm min} - \mu_\bff(\bx)) \Phi(Z) + \sigma_\bff(\bx) \phi(Z)
\end{align}
 where $Z \equiv (y_{\rm min} - \mu_\bff(\bx) ) / \sigma_\bff(\bx)$
and where $\Phi(\cdot)$ and $\phi(\cdot)$ are the CDF and PDF of the unit Normal distribution,
respectively.
By maximizing Eqn.~\eqref{eqn:eiclosedform} over $\DD$ we can find query points
$\bx$ that balance exploration and exploitation.

\section{Bayesian Optimization with Sparse Axis-Aligned Subspaces}
\label{sec:method}
We now introduce the surrogate model we use for high-dimensional BO.
For a large number of dimensions, the space of functions mapping
$\DD$ to $\mathbb{R}$ is---to put it mildly---very large, even assuming a certain degree of smoothness.
To facilitate sample-efficient BO it is necessary to make additional assumptions.
Intuitively, we would like to assume that the dimensions of $\bx \in \DD$ exhibit a hierarchy of relevance.
For example in a particular problem we might have that $\{ \bx_{3}, \bx_{52} \}$ are crucial features
for mapping the principal variation of $\fobj$, $\{ \bx_{7}, \bx_{14}, \bx_{31}, \bx_{72} \}$ are of moderate importance,
while the remaining features are of marginal importance.
This motivates the following desiderata for our function prior:
\begin{enumerate}[topsep=0pt,itemsep=-0.75ex,partopsep=1ex,parsep=1ex]
\item Assumes a hierarchy of feature relevances
\item Encompasses a flexible class of smooth non-linear functions
\item Admits tractable (approximate) inference
\end{enumerate}

\subsection{\priorname{} Function Prior}
\label{sec:funcprior}
\begin{algorithm*}
    \DontPrintSemicolon 
    \KwIn{Objective function $\fobj$;
          initial evaluation budget $m \ge2$;
          total evaluation budget $T>m$;
          hyperparameter $\alpha$;
          number of NUTS samples $L$;
          and initial query set $\bx_{1:m}$ and evaluations $y_{1:m}$ (\emph{optional})
         }
    \KwOut{Approximate minimizer and minimum $(\bx_{\rm min}, y_{\rm min})$}
    If $\{ \bx_{1:m}, y_{1:m} \}$ is not provided, let $\bx_{1:m}$ be a Sobol sequence
        in $\DD$ and let $y_t = \fobj(\bx_t)$ for $t=1,...,m$. \\
    \For{$t = m+1, ..., T$} {
        Let $\mathcal{H}_t = \{\bx_{1:t-1}, y_{1:t-1}\}$ and $y_{\rm min}^{\;t} = {\rm min}_{s< t} y_s$. \\
        Fit \priorname{} GP in Eqn.~\eqref{eqn:prior} to $\mathcal{H}_t$ using NUTS to obtain $L$
              hyperparameter samples $\{\psi_\ell^t \}$. \\
      Optimize the expected improvement in Eqn.~\eqref{eqn:eiclosedform2} to obtain
            $\bx_t = {\rm argmax}_{\bx} \; \rm{EI}(\bx |  y_{\rm min}^{\;t}, \{\psi_\ell^t \})$. \\
      Query $\fobj$ and set $y_t = \fobj(\bx_t)$.
    }
        \Return{$(\bx_{\rm min}, y_{\rm min})$} where $(\bx_{\rm min}, y_{\rm min}) \equiv (\bx_{{t_{\rm min}}}, y_{t_{\rm min}})$ and $t_{\rm min} = {\rm argmin}_{t} y_t $.
        \caption{We outline the main steps in \algoname{} when NUTS is used for inference.
        To instead use MAP we simply swap out line 4. For details on inference see Sec.~\ref{sec:inference};
        for details on EI maximization see Sec.~\ref{sec:acquisitionstrat}.}
    \label{algo}
    \end{algorithm*}

To satisfy our desiderata we introduce a GP model with a structured prior over the kernel hyperparameters, in particular one that
induces sparse structure in the (inverse squared) length scales $\rho_i$. In detail we define the following model:
\begin{align}
\label{eqn:prior}
&[{\rm kernel \; variance}] \;\;\;\;                 &&\sigma_k^2 \sim \mathcal{LN}(0, 10^2)  \\
&[{\rm global \; shrinkage}] \;\;\;\;      &&\tau \sim  \mathcal{HC}(\alpha)  \nonumber   \\
&[{\rm length \; scales}] \;\;\;\;              && \rho_i \sim  \mathcal{HC}(\tau)  \;\;\;\;\;\;\;\;\; {\rm for} \;\;  i=1,...,D.  \nonumber  \\
&[{\rm function \; values}] \;\;\;\;               &&\bff\sim \NN(\bm{0}, K^{\psi}_{\bX\bX})  \;\;\;
                                                               {\rm with} \;\; \psi = \{\rho_{1:d}, \sigma_k^2\}      \nonumber \\
&[{\rm observations}] \;\;\;\;              && \by \sim \NN(\bff, \sigma^2 \mathbb{1}_N)   \nonumber
\end{align}
where $\mathcal{LN}$ denotes the log-Normal distribution and $\mathcal{HC}(\alpha)$ denotes the half-Cauchy distribution,
i.e.~$p(\tau | \alpha) \propto (\alpha^2 + \tau^2)^{-1}\mathbb{1}(\tau > 0)$, and
$p(\rho_i | \tau) \propto (\tau^2 + \rho_i^2)^{-1}\mathbb{1}(\rho_i > 0)$.
Here $\alpha>0$ is a hyperparameter that controls the level of shrinkage (our default is $\alpha=0.1$).
We use an RBF kernel, although other choices like the Mat\'ern-5/2 kernel are also possible.
We also set $\sigma^2 \to 10^{-6}$, since we focus on noise-free objective functions $\fobj$.
Noisy objective functions can be accommodated by placing a weak prior on $\sigma^2$, for example $\sigma^2 \sim \mathcal{LN}(0, 10^2)$.

The \priorname{} function prior defined in \eqref{eqn:prior} has the following important properties.
First, the prior on the kernel variance $\sigma_k^2$ is weak (i.e.~non-informative).
Second, the level of global shrinkage (i.e.~sparsity)
is controlled by the scalar $\tau>0$, which tends to concentrate near zero due to the half-Cauchy prior.
Third, the (inverse squared) length scales $\rho_i$ are also governed by half-Cauchy priors, and thus they too tend to concentrate near zero (more precisely for most $i$ we expect $\rho_i \lesssim \tau$).
Consequently most of the dimensions are `turned off' in accord with the principle of \emph{automatic relevance determination} introduced by \cite{mackay1994automatic}. Finally, while the half-Cauchy priors favor values near zero, they have heavy tails.
This means that if there is sufficient evidence in the observations $\by$, the posterior over $\tau$ will be pushed to higher values, thus reducing the level of shrinkage and allowing more of the $\rho_i$ to escape zero, effectively `turning on' more dimensions.
The parsimony inherent in our function prior is thus \emph{adaptive}: as more data is accumulated, more of the $\rho_i$ will escape zero, and posterior mass will give support to a richer class of functions.
This is in contrast to a standard GP fit with maximum likelihood estimation (MLE), which will generally exhibit non-negligible $\rho_i$ for most dimensions---since there is no mechanism regularizing the length scales---typically resulting in drastic overfitting in high-dimensional settings.

Conceptually, our function prior describes functions defined on sparse axis-aligned subspaces, thus the name of our prior (\priorname) and our method (\algoname).

\subsection{Inference}
\label{sec:inference}

Doing inference for the model defined in Sec.~\ref{sec:funcprior} is challenging because of the dimension of the
latent space and the many non-linearities.
Thankfully, the latent variables in our model are continuous (and the joint density is differentiable), so we can leverage efficient gradient-based inference techniques.
In this section we describe the two inference strategies we pursue.
The first relies on the No-U-Turn sampler (NUTS) \citep{hoffman2014no},
an adaptive variant of Hamiltonian Monte Carlo that is the gold standard for inference in models like ours.
The second is a maximum a posteriori (MAP) approach, which trades off fidelity of the posterior approximation
for faster runtime.
In both cases we make use of the marginal likelihood $p(\by|\bX, \psi)$ in Eqn.~\eqref{eqn:mll}, i.e.~we integrate out the
latent function $\bff$ analytically.

\subsubsection{No-U-Turn Sampler (NUTS)}
\label{sec:nuts}

We use the NUTS sampler implemented in NumPyro \citep{phan2019composable} to
target the un-normalized joint density
\begin{align}
\label{eqn:unnormalized}
 p(\by|\bX, \psi) p(\psi | \tau) p(\tau) \propto p(\tau, \psi | \bX, \by).
\end{align}
Here $p(\psi | \tau) p(\tau)$ denotes the density over the kernel hyperparameters
$\psi$ and shrinkage parameter $\tau$ given in Eqn.~\eqref{eqn:prior}.
After running NUTS we obtain $L$ approximate posterior samples for the kernel hyperparameters, $\{\psi_\ell \}_{\ell=1}^L$.
The cost of obtaining a posterior sample is $\OO(N^3D)$ where $N$ is the total number of datapoints
and $D$ is the dimension of the input domain $\DD$.\footnote{The factor of $D$ comes from computing
terms that arise in the gradients of Eqn.~\eqref{eqn:mll}.}
Thus our method inherits the scalability bottleneck of all BO methods that rely on GPs and is
most suitable for moderate numbers of datapoints, e.g.~$N \lesssim 500$.
The kernel hyperparameters can then be plugged into the closed form GP predictive formulae
in Eqn.~\eqref{eqn:gppredmean}-\eqref{eqn:gppredvar}.

\subsubsection{Maximum a posteriori (MAP)}
\label{sec:map}

In MAP we target the same un-normalized density as in Eqn.~\eqref{eqn:unnormalized}, with a few small differences.
First, since MAP is formulated as an optimization problem w.r.t.~the target density, the result
of inference is a single point estimate and not a bag of samples as in NUTS.
Second, we remove the prior over $\tau$ and instead learn separate models for
a small number $S$ of pre-selected\footnote{Note that this means that
\algoname-\texttt{MAP} does not require specifying the hyperparameter $\alpha$.} values of $\tau$, e.g.~$\tau_s \in \{10^{-1},10^{-2},10^{-3}\}$.
Thus after convergence 
we obtain $S$ point estimates $\{(\psi_s)\}_{s=1}^S$.
Finally, to choose between these $S$ point estimates we use a leave-one-out measure of the predictive log likelihood to select the best performing $\psi_s$.
See Sec.~\ref{sec:suppmap} in the supplementary materials for details.

\subsection{Acquisition Strategy}
\label{sec:acquisitionstrat}
We use expected improvement (EI) as our acquisition function given its simplicity,
favorable computational properties, and good empirical performance.
We begin by noting that the expression for EI given
in Eqn.~\eqref{eqn:eiclosedform} depends on the kernel hyperparameters $\psi$ through $\mu_\bff(\bx)$ and $\sigma_\bff^2(\bx)$.
Thus in our context where $\psi$ is a latent variable, the expected improvement is defined by averaging Eqn.~\eqref{eqn:eiclosedform}
over posterior samples $\{ \psi_\ell \}_{\ell}^L \sim p(\psi | \mathcal{H}) $
\begin{align}
\label{eqn:eiclosedform2}
\rm{EI}(\bx |  y_{\rm min}, \{ \psi_\ell \}) \equiv \frac{1}{L} \sum_{\ell=1}^L \rm{EI}(\bx |  y_{\rm min}, \psi_\ell)
\end{align}
where in Eqn.~\eqref{eqn:eiclosedform2} we assume we have obtained $L$ samples from NUTS.

An essential property of Eqn.~\eqref{eqn:eiclosedform2} is that it is differentiable w.r.t.~$\bx$ and thus
can be efficiently optimized with gradient methods.
In practice we optimize Eqn.~\eqref{eqn:eiclosedform2} by generating a Sobol sequence in $\DD$ to find a small
number $K$ of promising starting points $\{\tilde{\bx}_k \}_{k=1}^K$ and then use these to initialize $K$ runs
of L-BFGS-B to obtain the query point
\begin{align}
\bx_{\rm next} = {\rm argmax}_\bx {\rm EI}(\bx |  y_{\rm min}, \{ \psi_\ell \})
\end{align}
See the supplementary materials for further details and Alg.~\ref{algo} for a complete outline of the \algoname{} algorithm.

\subsection{Discussion}
\label{sec:methoddisc}

We note that the axis-aligned structure of our model need not be as restrictive as one might at first assume.
For example, suppose that $\fobj$ can be written as $\fobj(\bx) = g(\bx_3 - \bx_7)$ for some $g: \mathbb{R} \to  \mathbb{R}$.
In order for our model to capture the structure of $\fobj$, both $\bx_3$ and $\bx_7$ need to be identified
as relevant. 
In many cases we expect this to be possible with a relatively small number of samples. While it is true that
identifying the direction $\bz = \bx_3 - \bx_7$ could be even easier in a different coordinate system, inferring
non-axis-aligned subspaces would come at the cost of substantially increased computational cost. More importantly,
by searching over a much larger set of subspaces our surrogate model would likely be much more susceptible to overfitting.
Given that for many problems we expect much of the function variation to be captured by axis-aligned blocks
of input features, we view our axis-aligned assumption as a good compromise between flexibility and parsimony.
Indeed in Sec.~\ref{sec:rotated} in the supplementary materials we describe an experiment in which we construct
objective functions with significant non-axis-aligned structure by performing a random rotation on an objective function
that is axis-aligned.
We find that SAASBO performs well in this challenging setting; see Fig.~\ref{fig:rotated_hartmann} in the supplementary material.
Importantly, our modeling approach does not sacrifice any of the many benefits of GPs (e.g.~flexible non-linearity
and non-parametric latent functions) nor do we need to make any unduly strong assumptions about $\fobj$ (e.g.~additive decomposition).

It is important to emphasize that it is by design that the model defined in Sec.~\ref{sec:funcprior} does not include any discrete latent variables.
A natural alternative to our model would introduce $D$ binary-valued variables that control whether or not a given dimension is relevant to modeling $\fobj$.
However, inference in any such model is very challenging as it requires exploring a discrete space of size $2^D$.
Our model can be understood as a continuous relaxation of such an approach.
Indeed, the structure of our sparsity-inducing prior closely mirrors the justly famous
Horseshoe prior \citep{carvalho2009handling}, which is a popular prior for Sparse Bayesian linear regression.
We note that in contrast to the linear regression setting of the Horseshoe prior,
our sparsity-inducing prior governs inverse squared length scales in a non-linear kernel and \emph{not} variances.
We discuss this point in more detail in Sec.~\ref{sec:nodisc} in the supplementary materials.

\section{Experiments}
\label{sec:exp}
We present an empirical validation of our approach.
In Sec.~\ref{sec:fit}-\ref{sec:relevance} we characterize the behavior of \algoname{} in controlled settings.
In Sec.~\ref{sec:sec_synthetic}-\ref{sec:sec_mopta08} we benchmark \algoname{} against a number of state-of-the-art methods for high-dimensional BO.
An open source implementation of \algoname{} that relies on Pyro \citep{bingham2019pyro} will be made
available in BoTorch \citep{balandat2019botorch}, while a NumPyro \citep{phan2019composable} version is available at \url{https://github.com/martinjankowiak/saasbo}.

\subsection{The \priorname{} Prior Provides Good Model Fit in High Dimensions}
\label{sec:fit}

\begin{figure}[!ht]
    \includegraphics[width=0.48\textwidth]{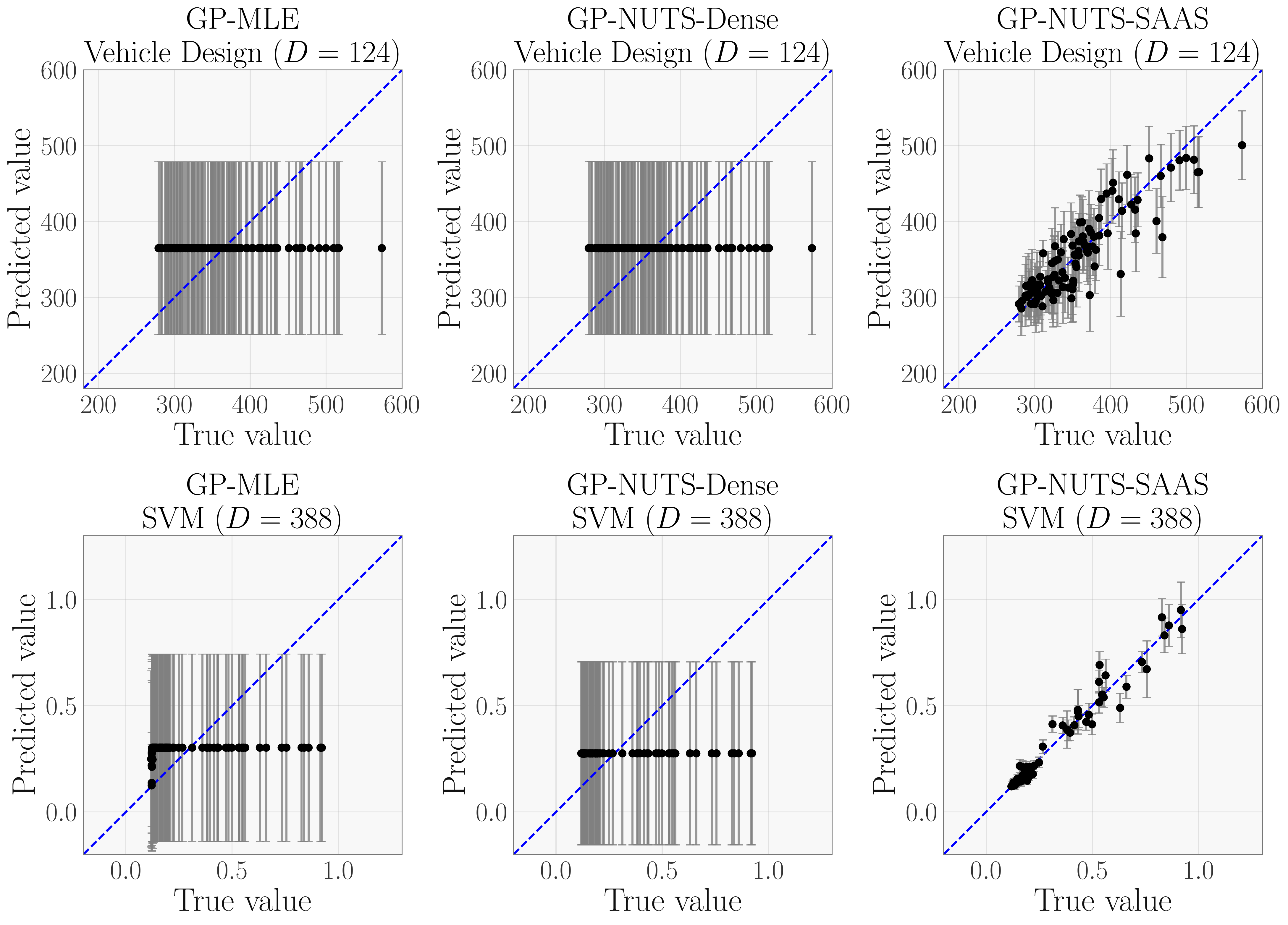}
    \caption{
      We compare model fit for three models on the $D=124$ vehicle design problem and the $D=388$ SVM problem
      (see Sec.~\ref{sec:sec_svm}-\ref{sec:sec_mopta08} for details).
    We compare:
    (left) a GP fit with MLE;
    (middle) a GP with weak priors fit with NUTS;
    and (right) a GP with a \priorname{} prior (this paper; see Eqn.~\eqref{eqn:prior}) fit with NUTS.
    For the vehicle design problem we use $100$ training points and for the SVM problem we use $50$ training points.
    We use $100$ test points for both problems.
    Only \priorname{} provides a good fit.
    In each figure mean predictions are depicted with dots and bars denote $95$\% confidence intervals.
    }
    \label{fig:model_cv}
\end{figure}

In Fig.~\ref{fig:model_cv} we demonstrate the importance of using a sparsity-inducing prior
like \priorname{} when fitting a GP in a high-dimensional domain.
For these high-dimensional problems, both maximum likelihood
estimation and full Bayesian inference for a GP with weak log-Normal
priors on the squared length scales $\rho_i^{-1}$ concentrate on
solutions in which the vast majority of the $\rho_i$ are $\OO(1)$. Consequently
with high probability the kernel similarity between a randomly chosen test point and any of the
$N=100$ training data points is $\OO (\exp(-D)) \approx 0$, with the result that both
these models revert to a trivial mean prediction across most of the domain.
By contrast, the \priorname{} prior only allows a few $\rho_i$ to escape zero, resulting
in a model that is much more useful for exploration and exploitation of the most important design variables.

\subsection{\algoname{} can quickly identify the most relevant dimensions}
\label{sec:relevance}
\begin{figure*}[!ht]
    \centering
    \includegraphics[width=0.99\textwidth]{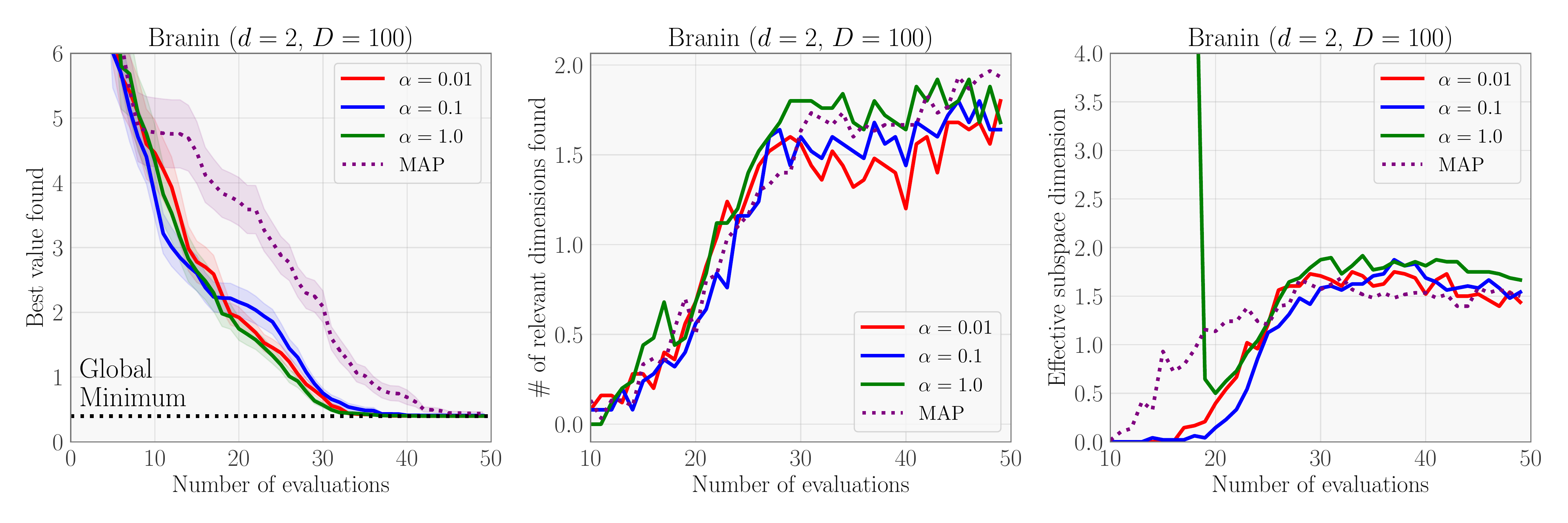}
    \caption{We explore how \algoname{} performs on Branin ($D=100$), comparing
    \algoname-\texttt{NUTS} for three values of the sparsity controlling
    hyperparameter $\alpha$ to \algoname-\texttt{MAP}.
    Each curve corresponds to $60$ independent replications of Algorithm \ref{algo}.
    {\bf Left}: We compare performance w.r.t.~the best minimum found (the mean is depicted by a thick line and
    shaded bands denote standard errors).
    {\bf Middle}: We depict the mean number of \emph{relevant} dimensions found, where
    a relevant dimension is declared `found' if its corresponding $\rm{PosteriorMedian}(\rho_k)$
    is among the two largest $\{ \rm{PosteriorMedian}(\rho_i) \}_{i=1}^D$.
    {\bf Right}: We depict the mean effective subspace dimension, defined to be the number of dimensions
    for which ${\rm{PosteriorMedian}(\rho_k) > 0.5}$.}
    \label{fig:branin3alphas}
\end{figure*}

We characterize the behavior
of \algoname{} in a controlled setting where we embed the two-dimensional Branin function in $D=100$ dimensions.
First, we explore the degree to which \algoname's performance depends on the approximate inference algorithm used,
in particular comparing NUTS to MAP (see Sec.~\ref{sec:inference} for details on inference).
In Fig.~\ref{fig:branin3alphas} (left) we see that NUTS outperforms MAP by a considerable margin.
In Fig.~\ref{fig:branin3alphas} (middle and right) we demonstrate that both inference methods
are able to reliably identify the two relevant dimensions after $\sim 20-30$ evaluations.

Why does NUTS outperform MAP even though MAP is able to identify the relevant subspace?
We hypothesize that the primary reason for the superior performance of NUTS is that the EI objective
in Eqn.~\eqref{eqn:eiclosedform2} is considerably more robust when averaged over multiple samples of the
GP kernel hyperparameters. In particular, averaging over multiple samples---potentially from
distinct modes of the posterior---appears to mitigate EI's tendency to seek out the boundary of the domain $\DD$.
For this reason we use NUTS for the experiments in this work, noting that while we obtain good performance with MAP in some problem
settings we find that NUTS is significantly more robust.

Next, we explore the dependence of \algoname-\texttt{NUTS} on the hyperparameter $\alpha$.
In Fig.~\ref{fig:branin3alphas} (left) we see that there is
minimal dependence on $\alpha$, with the three values leading to similar optimization performance.
In Fig.~\ref{fig:branin3alphas} (middle and right) we see that, as expected, smaller values of
$\alpha$ are more conservative (i.e.,~prefer smaller subspaces), while larger values of $\alpha$
are less conservative (i.e.,~prefer larger subspaces). We note, however, that this effect is most pronounced
when only a small number of datapoints have been collected. After $\sim 20$ function evaluations
the observations overwhelm the prior $p(\tau)$ and the posterior quickly concentrates on the two relevant dimensions.

Given the good performance of all three values of $\alpha$, for the remainder of our experiments we
choose the intermediate value $\alpha=0.1$.
While performance can perhaps be improved
in some cases by tuning $\alpha$, we find it encouraging that we can get good performance with
a single $\alpha$. We emphasize that $\alpha$ is the only hyperparameter that governs the function prior, and that all remaining hyperparameters
control the computational budget (e.g.~the number of NUTS samples $L$).
This is in contrast to the many methods for high-dimensional BO that rely on several (potentially sensitive)
hyperparameters such as the dimension $d_e$ of a random embedding.

\begin{figure*}[!ht]
    \centering
    \includegraphics[width=0.99\linewidth]{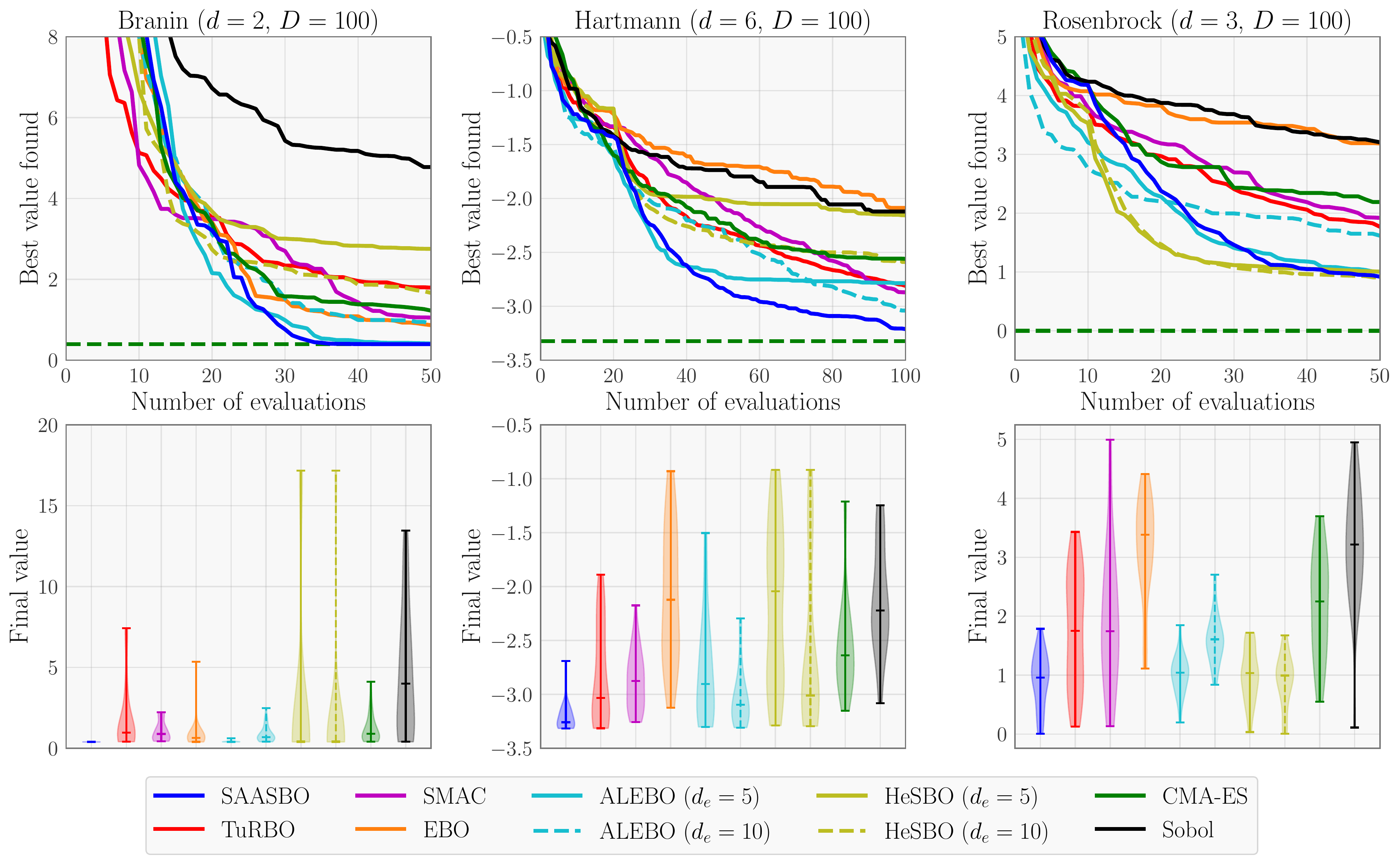}
      \caption{We compare \algoname{} to seven baseline methods on three $d-$dimensional functions embedded
      in $D=100$ dimensions. In each case we do $30$ independent replications.
      {\bf Top row:} For each method we depict the mean value of the best minimum found at a given iteration.
      {\bf Bottom row:} For each method we depict the distribution over the final approximate minimum $y_{\rm min}$ encoded as
      a violin plot, with horizontal bars corresponding to $5$\%, $50$\%, and $95$\% quantiles.
      }
    \label{fig:synthetic}
  \end{figure*}

\subsection{Baselines}
We compare \algoname{} to a comprehensive selection of baselines: ALEBO, CMA-ES, EBO, HeSBO, SMAC, Sobol, and TuRBO.
ALEBO~\citep{letham2020re} is chosen as a representative random embedding method, as it improves upon the original REMBO method~\citep{wang2016bayesian}.
Additionally, we compare to HeSBO, which uses hashing and sketching to project low-dimensional points up to the original space~\citep{nayebi2019framework}.
The EBO method by~\citet{wang2018batched} exploits additive structure to scale to high-dimensional spaces.
We also compare to CMA-ES \citep{hansen2003reducing}, which is a popular evolutionary method that is often competitive with BO methods on high-dimensional problems, see e.g.,~\citep{letham2020re}.
TuRBO~\citep{turbo} uses a trust region centered at the best solution to avoid exploring highly uncertain parts of the search space.
We also include an additional BO method that does not rely on GPs, namely SMAC~\citep{hutter2011sequential}.
Finally, we also compare to scrambled Sobol sequences~\citep{owen2003quasi}.

We use the default settings for all baselines.
For ALEBO and HeSBO we evaluate both $d_e=5$ and $d_e=10$ on the three synthetic problems in Sec.~\ref{sec:sec_synthetic}.
As $d_e=5$ does not perform well on the three real-world applications in Sec.~\ref{sec:sec_rover}-\ref{sec:sec_mopta08}, we instead evaluate $d_e=10$ and $d_e=20$ on these problems.

We also mention a baseline method for which we do not report results, since it underperforms random search.
Namely for our surrogate model we use a quadratic polynomial over $\DD$ with $\OO(D^2)$ coefficients governed by a
sparsity-inducing Horseshoe prior \citep{carvalho2009handling}.
As in \cite{baptista2018bayesian}, this finite feature expansion admits efficient inference with a Gibbs sampler.
Unfortunately, in our setting, where $\DD$ is continuous and not discrete, this leads to pathological behavior
when combined with EI, since the minima of simple parametric models are very likely to be found at the boundary of $\DD$.
This is in contrast to the mean-reverting behavior of a GP with a RBF or Mat\'ern kernel, which is a much more appropriate modeling assumption in high dimensions.

\begin{figure*}[!ht]
    \centering
    \includegraphics[width=0.99\linewidth]{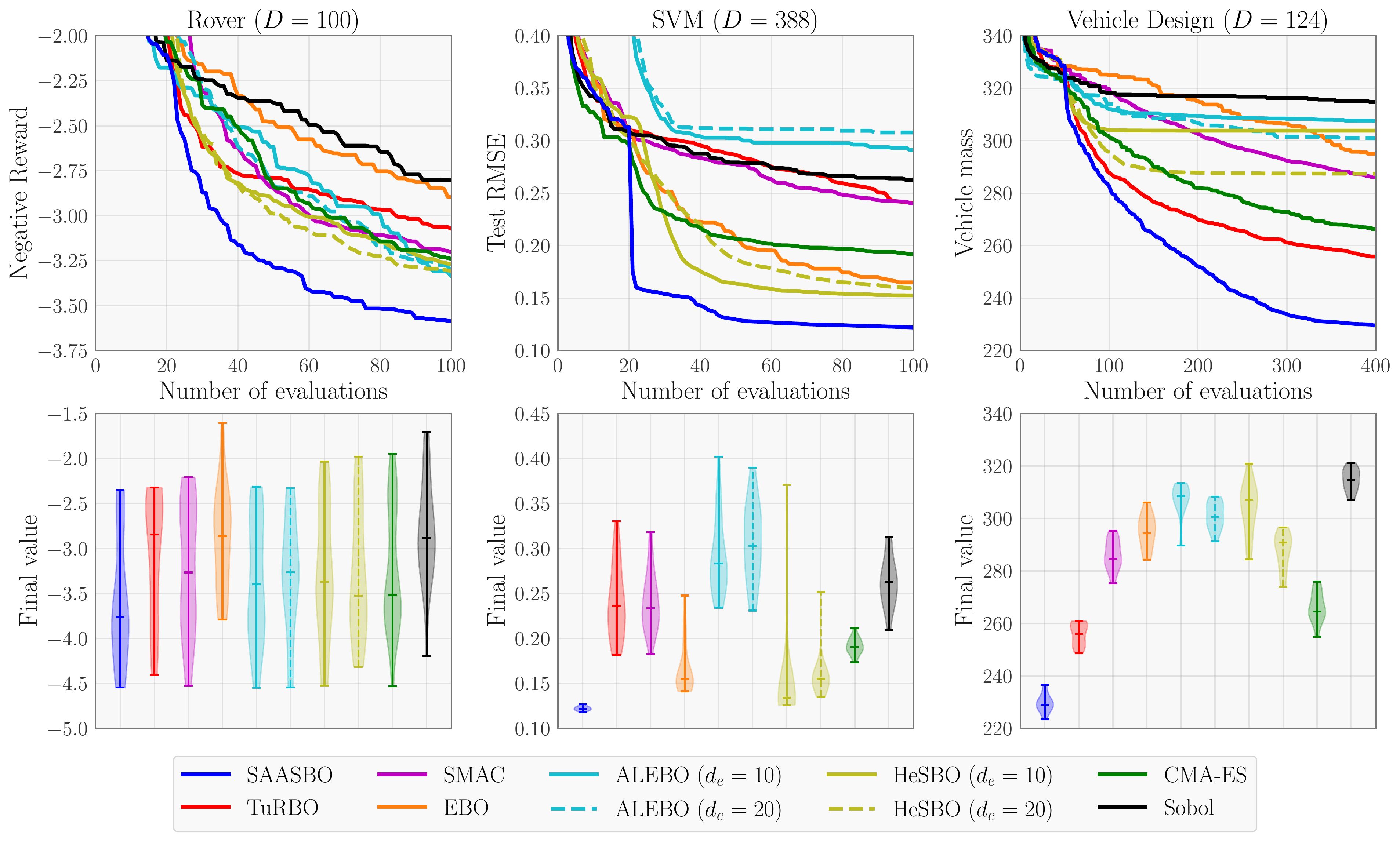}
      \caption{We compare \algoname{} to baseline methods on rover trajectory planning ($D=100$),
               SVM hyperparameter tuning ($D=388$), and MOPTA vehicle design ($D=124$).
               We do $30$ independent replications for Rover and SVM and $15$ replications for MOPTA.
      {\bf Top row:} For each method we depict the mean value of the best minimum found at a given iteration.
      {\bf Bottom row:} For each method we depict the distribution over the final approximate minimum $y_{\rm min}$ encoded as
      a violin plot, with horizontal bars corresponding to $5$\%, $50$\%, and $95$\% quantiles.
    }
    \label{fig:real_world}
  \end{figure*}

\subsection{Synthetic problems}
\label{sec:sec_synthetic}

In this section we consider the Branin ($d=2$), Hartmann ($d=6$), and Rosenbrock ($d=3$)
test functions embedded in a $D=100$ space.\footnote{That is to say each synthetic function depends on exactly $d$ variables and is independent of the remaining $D-d$ variables.}
These are problems with unambiguous low-dimensional structure where we expect both random embedding methods and \algoname{} to
perform well.

Fig.~\ref{fig:synthetic} shows that \algoname{} and ALEBO-$5$ perform the best on Branin.
\algoname{} performs the best on Hartmann followed by ALEBO-$10$.
HeSBO performs well on Rosenbrock and the final performance of \algoname, HeSBO-$5$, HeSBO-$10$, and ALEBO-$5$ are similar.
However, both ALEBO and HeSBO show significant sensitivity to the embedded subspace dimension on at least two of the three problems, highlighting a serious downside of random embedding methods. Crucially this important hyperparameter needs to be chosen before the start of optimization and is not learned.

\subsection{Rover trajectory planning}
\label{sec:sec_rover}
We consider a variation of the rover trajectory planning problem from~\citep{wang2018batched} where the task is to find an optimal trajectory through a 2d-environment.
In the original problem, the trajectory is determined by fitting a B-spline to $30$ waypoints and the goal is to optimize the locations of these waypoints.
This is a challenging problem that requires thousands of evaluations to find good solutions, see e.g.~\citep{turbo}.
To make the problem more suitable for small evaluation budgets, we require that the B-spline starts and ends at the pre-determined starting position and destination.
We also increase the dimensionality to $D=100$ by using $50$ waypoints.
Fig.~\ref{fig:real_world} shows that \algoname{} performs the best on this problem.
This problem is challenging for all methods, each of which had at least one replication where the final reward was below 2.5.\looseness-1

\subsection{Hyperparameter tuning of an SVM}
\label{sec:sec_svm}
We define a hyperparameter tuning problem using a kernel support vector machine (SVM) trained
on a $385$-dimensional regression dataset.
This results in a $D=388$ problem, with $3$ regularization parameters and $385$
kernel length scales. We expect this problem to have some amount of low-dimensional structure,
as we expect the regularization parameters to be most relevant, with a number of
length scales of secondary, but non-negligible importance. This intuition
is confirmed in Fig.~\ref{fig:svm388relevance} in the supplementary materials, which demonstrates
that \algoname{} quickly focuses on the regularization parameters, explaining the superior performance of \algoname{} seen in Fig.~\ref{fig:real_world}.
ALEBO makes little progress after iteration $30$, indicating that there may not be any good solutions within the random embeddings.
HeSBO and EBO do better than the other methods, but fail to match the final performance of \algoname.

\subsection{Vehicle design}
\label{sec:sec_mopta08}
We consider the vehicle design problem MOPTA08, a challenging real-world high-dimensional BO problem \citep{jones2008large}.
The goal is to minimize the mass of a vehicle subject to $68$ performance constraints.
The $D=124$ design variables describe materials, gauges, and vehicle shape.
To accommodate our baseline methods,
While some methods such as Scalable Constrained Bayesian Optimization (SCBO)~\citep{eriksson2020scalable} can handle this constrained problem with thousands of evaluations,
we convert the hard constraints into a soft penalty, yielding a scalar objective function.
Fig.~\ref{fig:real_world} shows that \algoname{} outperforms other methods by a large margin.
TuRBO and CMA-ES perform better than the remaining methods, which fail to identify good solutions.
While this problem does not have obvious low-dimensional structure, our flexible \priorname{} prior still results in superior optimization performance.

\begin{figure}[!t]
  \centering
  \includegraphics[width=0.9\linewidth]{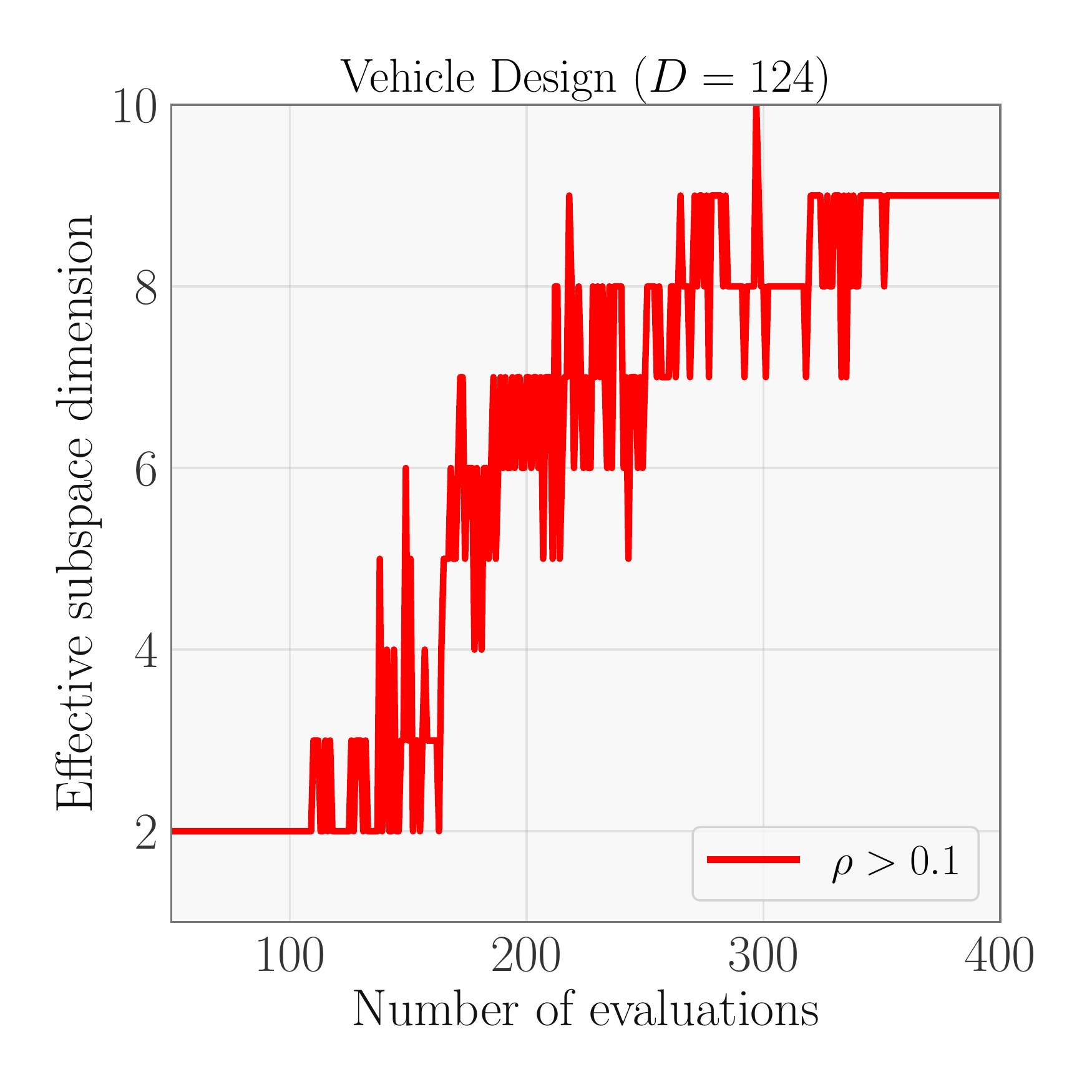}
  \caption{
  We depict the effective subspace dimension during a single run of Algorithm ~\ref{algo} on the MOPTA vehicle
    design problem. Here the effective
    subspace dimension is the number of dimensions
    for which ${\rm{PosteriorMedian}(\rho_k) > \xi}$, with $\xi=0.1$ an arbitrary cutoff.}
  \label{fig:mopta_relevance}
\end{figure}

In Fig.~\ref{fig:mopta_relevance} we see that during the course of a single run
of \algoname{} on this problem,
the effective dimension of the identified subspace steadily increases
from about $2$ to about $10$ as more evaluations are collected.
Using an increasingly flexible surrogate model over the course of
optimization is key to the excellent optimization performance of \algoname{}.

\section{Discussion}
\label{sec:discussion}
Black-box optimization in hundreds of dimensions presents
a number of challenges, many of which can be traced to the
many degrees of freedom that characterize high-dimensional spaces.
The majority of approaches to Bayesian optimization try to circumvent this
potential hazard
by reducing the effective dimensionality of the problem.
For example random projection methods like
ALEBO and HeSBO work directly in a low-dimensional space, while
methods like TuRBO or LineBO constrain the domain over which the acquisition
function is optimized. We take the view that it is much more natural to work directly
in the full space and instead rely on a sparsity-inducing function prior to
mitigate the curse of dimensionality.

As we have shown in a comprehensive set of experiments, \algoname{} outperforms state-of-the-art BO methods on several synthetic and real-world problems.
Our approach provides several distinct advantages: we highlight three.
First, it preserves---and therefore can exploit---structure in the input domain, in contrast
to methods like ALEBO or HeSBO which risk scrambling it.
Second, it is adaptive and exhibits little sensitivity to its hyperparameters.
Third, it can naturally accommodate both input and output constraints,
in contrast to methods that rely on random projections, for which input constraints
are particularly challenging.

While we have obtained strikingly good performance using a simple acquisition strategy, it is likely that making the most of our \priorname{} function prior will require a decision-theoretic framework that is better suited to high-dimensional settings.
This is an interesting direction for future elaborations of \algoname.

\begin{acknowledgements}
    We thank Neeraj Pradhan and Du Phan for help with NumPyro and Maximilian Balandat for providing feedback on a draft manuscript.
\end{acknowledgements}

\bibliography{ref}

\appendix

\section{Inference}

\subsection{NUTS}
\label{sec:suppnuts}
We use the NUTS sampler implemented in NumPyro \citep{phan2019composable, bingham2019pyro},
which leverages JAX for efficient hardware acceleration \citep{bradbury2020jax}.
In most of our experiments (see Sec.~\ref{sec:expdetails} for exceptions)
we run NUTS for $768 = 512 + 256$ steps
where the first $N_{\rm warmup} = 512$ samples are for burn-in and (diagonal) mass matrix adaptation (and thus discarded),
and where we retain every $16^{\rm th}$ sample among the final
$N_{\rm post}=256$ samples (i.e.~sample thinning),
yielding a total of $L=16$ approximate posterior samples.
It is these $L$ samples that are then used to compute Eqns.~\eqref{eqn:gppredmean}, \eqref{eqn:gppredvar}, \eqref{eqn:eiclosedform2}. We also limit the maximum tree depth in NUTS to $6$.

\begin{figure}[!ht]
  \centering
  \includegraphics[width=0.8\linewidth]{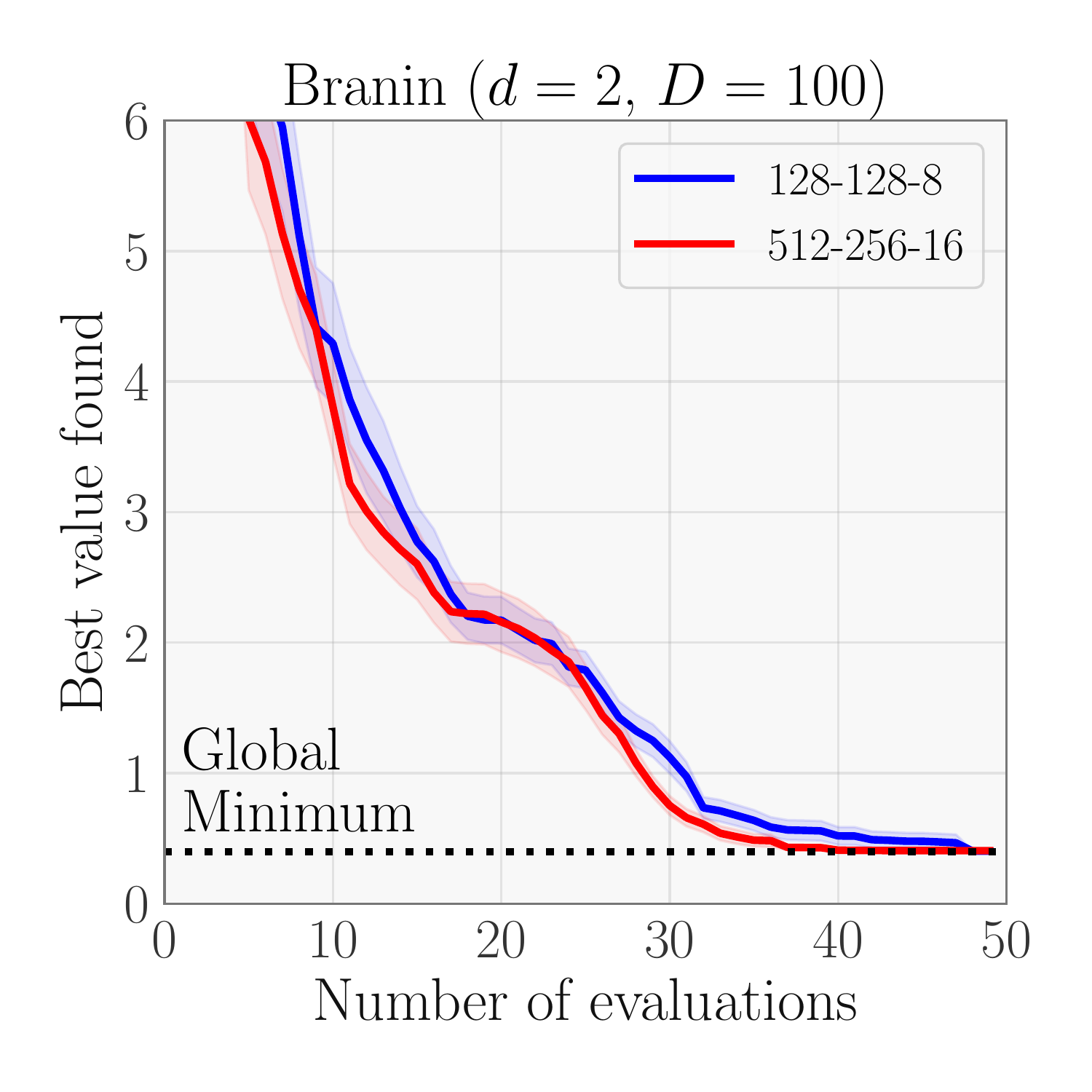}
    \caption{
        We depict how \algoname-\texttt{NUTS} performs on Branin
        as we reduce the sampling budget
        $(N_{\rm warmup}, N_{\rm post}, L) = (512, 256, 16)$
        to $(N_{\rm warmup}, N_{\rm post}, L) = (128, 128, 8)$.
    We compare performance w.r.t.~the best minimum found
    (the mean is depicted by a thick line and shaded bands denote standard errors).
    Each curve corresponds to $60$ independent replications of Algorithm \ref{algo}.
}
  \label{fig:fastnuts}
\end{figure}

We note that these choices are somewhat conservative, and in many settings we would expect good results with fewer samples. Indeed on the Branin test function, see Fig.~\ref{fig:fastnuts},
we find a relatively marginal drop in performance
when we reduce the NUTS sampling budget as follows:
i) reduce the number of warmup samples from $512$ to $128$;
ii) reduce the number of post-warmup samples from $256$ to $128$; and
iii) reduce the total number of retained samples from $16$ to $8$. We expect broadly
similar results for many other problems. See Sec.~\ref{sec:runtime} for corresponding runtime results.

It is worth emphasizing that while \algoname{} requires specifying a few hyperparameters
that control NUTS, these hyperparameters are purely computational in nature, i.e.~they
have no effect on the \priorname{} function prior.
Users simply choose a value of $L$ that meets their computational budget.
This is in contrast to e.g.~the embedding dimension $d_e$ that is required by ALEBO and
HeSBO: the value of $d_e$ often has significant effects on optimization performance.

To improve the geometry of the joint density defined by the model---and thus make NUTS
more efficient---we reparameterize the prior in Eqn.~\eqref{eqn:prior} as follows:
\begin{align}
&[{\rm global \; shrinkage}] \;\;\;\;      &&\tau \sim  \mathcal{HC}(\alpha)     \\
&[\rm reparameterized \; length \; scales] \;\;\;\;              && \tilde{\rho}_i \sim  \mathcal{HC}(1) \nonumber \\
&[{\rm effective \;length \; scales}] \;\;\;\;              && \rho_i = \tau \times \tilde{\rho}_i \nonumber
\end{align}
where we note that the final equation is a deterministic equality and HMC is performed in
the coordinate system defined by $\tilde{\rho}_i$. Note that this sort of reparameterization
can be implemented in NumPyro using the \texttt{deterministic} primitive.

We also note that it is possible to make \algoname-\texttt{NUTS} faster by means of the following modifications:
\begin{enumerate}
\item Warm-start mass adaptation with mass matrices from previous iterations.
\item Instead of fitting a new \priorname{} GP at each iteration, only fit
    every $M$ iterations (say $M=5$), and reuse hyperparameter samples $\{ \psi_\ell \}$
        across $M$ iterations of \algoname.
\end{enumerate}

\subsection{MAP}
\label{sec:suppmap}
We run the Adam optimizer \citep{kingma2014adam} for 1500 steps and with a learning
rate of $0.02$ and $\beta_1=0.50$  to maximize the log density
\begin{align}
U_s(\psi_s | \tau_s) = \log p(\by|\bX, \psi_s) + \log p(\psi_s | \tau_s)
\end{align}
w.r.t.~$\psi_s$
for $S=4$ pre-selected values of $\tau_s$: $\tau_s \in \{1, 10^{-1},10^{-2},10^{-3}\}$.
This optimization is trivially optimized across $S$.

For each $s=1,...,S$ we then compute the leave-one-out predictive log likelihood
using the mean and variance functions given in
Eqns.~\eqref{eqn:gppredmean}-\eqref{eqn:gppredvar}. We then choose the value
of $s$ that maximizes this predictive log likelihood and use the corresponding
kernel hyperparameter $\psi_s$ to compute the expected improvement
in Eqn.~\eqref{eqn:eiclosedform2}.

\subsection{No Discrete Latent Variables}
\label{sec:nodisc}
As discussed briefly in the main text, it is important
that the \priorname{} prior defined in Sec.~\ref{sec:funcprior} does not include any discrete latent variables.
Indeed a natural alternative to our model would introduce $D$ binary-valued
latent variables that control whether or not a given dimension is relevant to modeling $\fobj$.
However, inference in any such model can be very challenging, as it requires exploring an extremely large discrete space of size $2^D$.
Our model can be understood as a continuous relaxation of such an approach.
This is a significant advantage since it means we can leverage gradient information to efficiently explore the posterior.
Indeed, the structure of our sparsity-inducing prior closely mirrors the justly famous
Horseshoe prior \citep{carvalho2009handling}, which is a popular prior for Sparse Bayesian linear regression.
We note that in contrast to the linear regression setting of the Horseshoe prior,
our sparsity-inducing prior governs inverse squared length scales in a non-linear kernel and \emph{not} variances.
While we expect that any prior that concentrates $\rho_i$ at zero can exhibit good empirical performance in the setting
of high-dimensional BO, this raises the important question whether distributional assumptions other than those in Eqn.~\eqref{eqn:prior} may be better suited to governing our prior expectations about $\rho_i$.
Making a careful investigation of this point is an interesting direction for future work.

\section{Expected Improvement Maximization}
We first form a scrambled Sobol sequence $\bx_{1:Q}$ (see e.g.~\citep{owen2003quasi})
of length  $Q=5000$  in the $D$-dimensional domain $\DD$.
We then compute the expected improvement in Eqn.~\eqref{eqn:eiclosedform2}
in parallel for each point in the Sobol sequence. We then choose the top $K=3$
points in $\bx_{1:Q}$, that yield the largest EIs. For each of these $K$ approximate
maximizers
we run L-BFGS \citep{zhu1997algorithm} initialized with the approximate maximizer and
using the implementation provided by \texttt{Scipy}
(in particular \texttt{fmin\char`_l\char`_bfgs\char`_b})
to obtain the final query point $\bx_{\rm next}$, which (approximately) maximizes
Eqn.~\eqref{eqn:eiclosedform2}. We limit \texttt{fmin\char`_l\char`_bfgs\char`_b} to use a maximum
of 100 function evaluations.

\section{Runtime Experiment}
\label{sec:runtime}
We measure the runtime of \algoname{} as well as each baseline method
on the Branin test problem. See Table~\ref{tab:runtimes} for the results.
We record runtimes for both the default \algoname-\texttt{NUTS} settings described in
Sec.~\ref{sec:suppnuts} as well as one with a reduced NUTS sampling budget.
\begin{table}[!ht]
  \centering
  \caption{Average runtime per iteration on the Branin test function embedded in a $100$-dimensional space.
    Each method uses $m=10$ initial points and a total of $50$ function evaluations.
    Runtimes are obtained using a $2.4$ GHz $8$-Core Intel Core i9 CPU outfitted with $32$ GB of RAM.}
  \label{tab:runtimes}
  \begin{tabular}{c|c}
    Method & Time / iteration \\
    \hline
      \algoname \;(default) & $26.51$ seconds \\
      \algoname \;($128$-$128$-$8$) & $19.21$ seconds \\
    TuRBO & $1.52$ seconds \\
    SMAC & $12.12$ seconds \\
    EBO & $128.10$ seconds \\
    ALEBO ($d_e=5$) & $4.34$ seconds \\
    ALEBO ($d_e=10$) & $11.91$ seconds \\
    HeSBO ($d_e=5$) & $0.70$ seconds \\
    HeSBO ($d_e=10$) & $1.51$ seconds \\
    CMA-ES & $< 0.1$ seconds \\
    Sobol & $< 0.01$ seconds \\
  \end{tabular}
\end{table}
While \algoname{} requires more time per iteration than other methods such as TuRBO and HeSBO, the overhead is
relatively moderate in the setting where the black-box function $\fobj$ is very expensive to evaluate.
We note that after reducing the NUTS sampling budget to $(N_{\rm warmup}, N_{\rm post}, L) = (128, 128, 8)$
about $75$\% of the runtime is devoted to EI optimization. Since our current implementation
executes $K=3$ runs of L-BFGS serially, this runtime could be reduced further by executing L-BFGS in parallel.

\section{Additional Figures and Experiments}
\label{sec:addexp}

\subsection{Model fitting}
\begin{figure}[!ht]
  \includegraphics[width=0.48\textwidth]{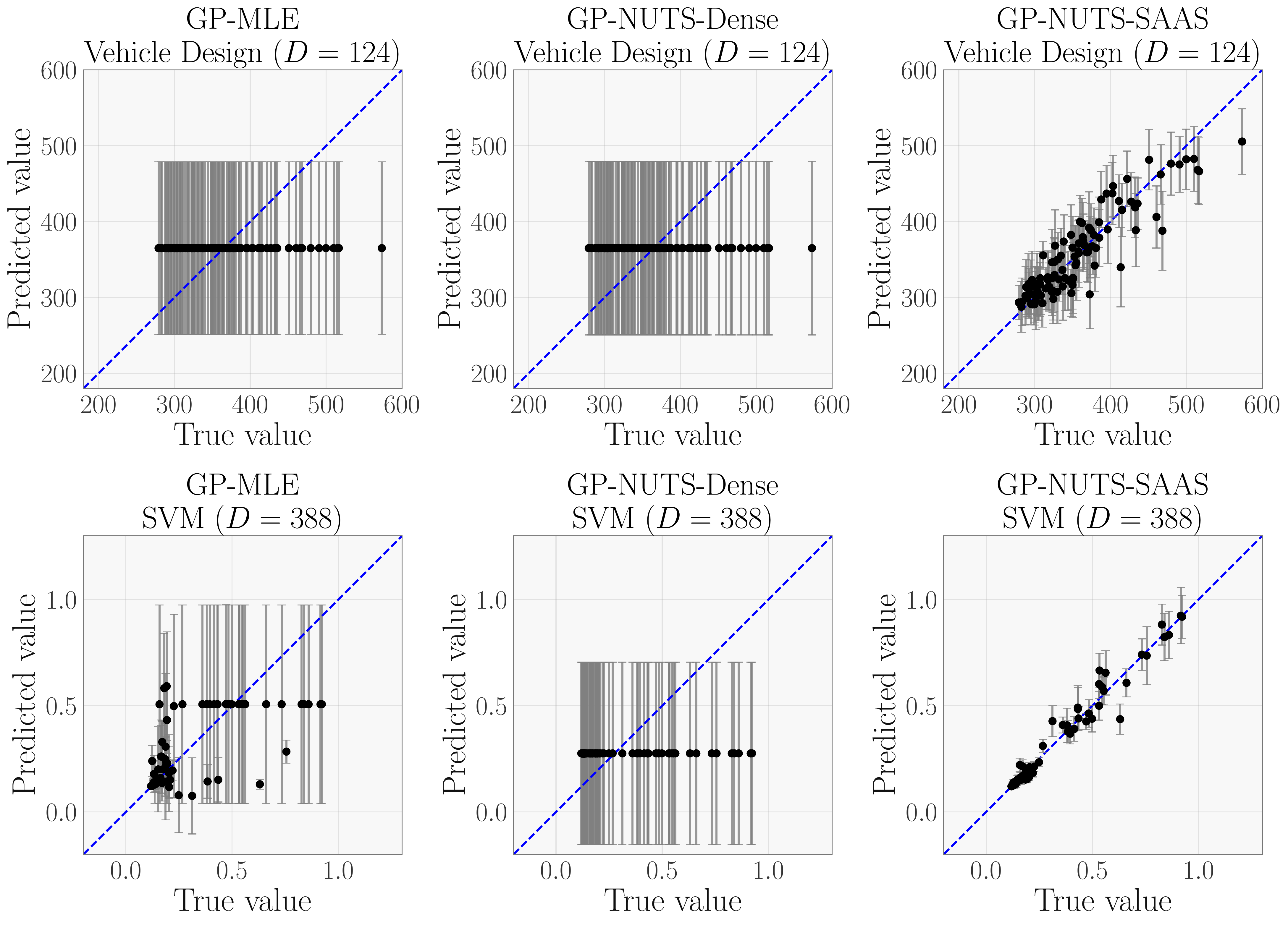}
  \caption{This figure is an exact reproduction of Fig.~\ref{fig:model_cv} in the main
  text apart from the use of a Mat\'ern-$5/2$ kernel instead of a RBF kernel.
  We compare:
  (left) a GP fit with MLE;
  (middle) a GP with weak priors fit with NUTS;
  and (right) a GP with a \priorname{} prior (this paper; see Eqn.~\eqref{eqn:prior}) fit with NUTS.
  For the vehicle design problem we use $100$ training points and for the SVM problem we use $50$ training points.
  We use $100$ test points for both problems.
  Only \priorname{} provides a good fit.
  In each figure mean predictions are depicted with dots and bars denote $95$\% confidence intervals.
  }
  \label{fig:model_cv_matern}
\end{figure}

In Fig.~\ref{fig:model_cv_matern} we reproduce the experiment described in
Sec.~\ref{sec:fit}, with the difference that we replace
the RBF kernel with a Mat\'ern-$5/2$ kernel.

\, \newline
We note that the qualitative behavior
in Fig.~\ref{fig:model_cv_matern} matches the behavior in Fig.~\ref{fig:model_cv}. In
particular, only the sparsity-inducing \priorname{} function prior
provides a good fit. This emphasizes that the potential for drastic overfitting that arises
when fitting a non-sparse GP in high dimensions is fundamental and is not ameliorated by
using a different kernel. In particular the fact that the Mat\'ern-$5/2$ kernel decays
less rapidly at large distances as compared to the RBF kernel (quadratically instead of
exponentially) does not prevent the non-sparse models from yielding essentially trivial predictions
across most of the domain $\DD$.

\subsection{SVM relevance plots}
\begin{figure}[!ht]
  \centering
  \includegraphics[width=0.99\linewidth]{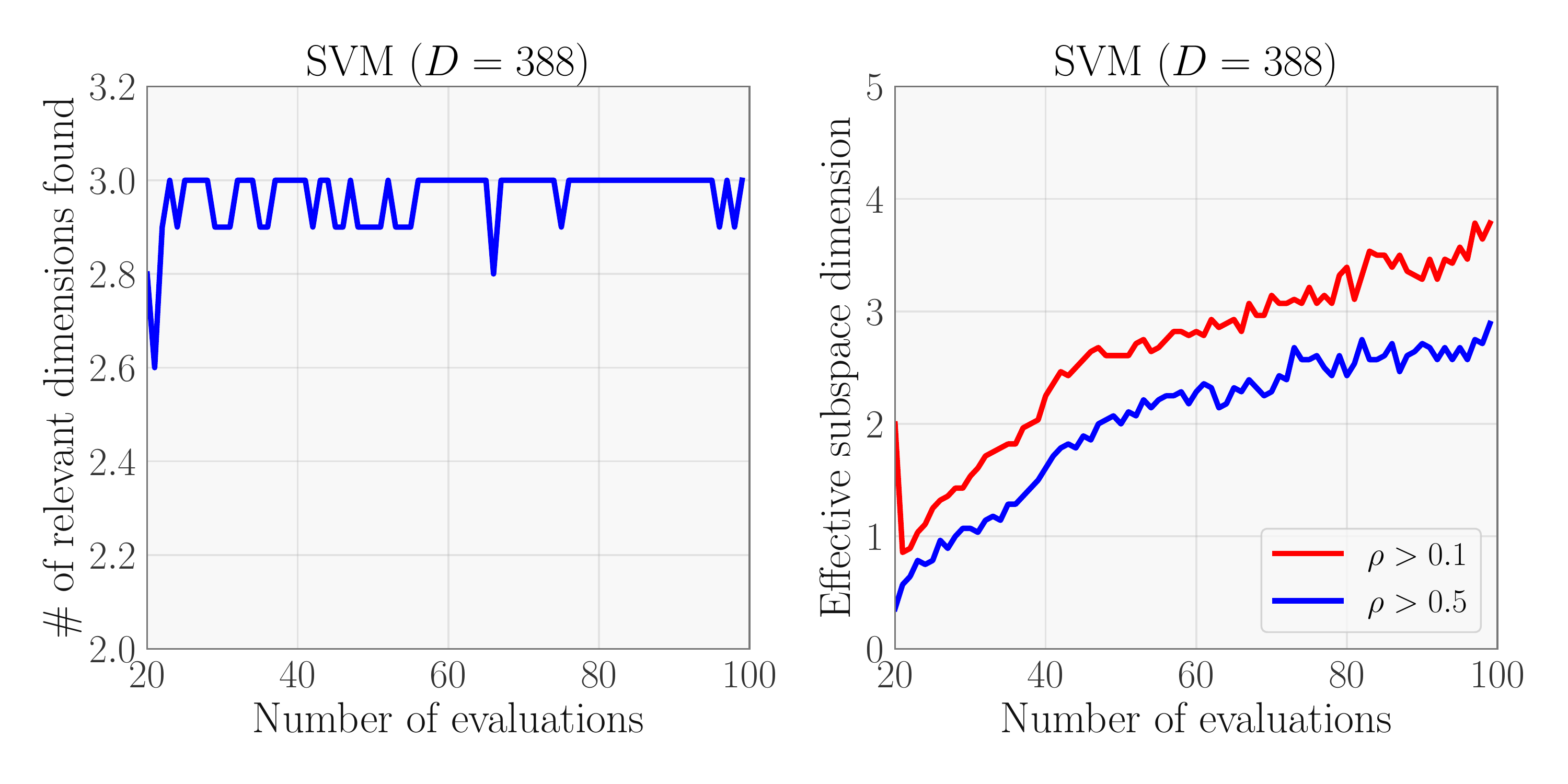}
  \caption{
  {\bf Left}: We depict the mean number of regularization hyperparameters that have been
  `found' in the SVM problem, where
  a regularization hyperparameter is `found' if its corresponding $\rm{PosteriorMedian}(\rho_k)$
  is among the three largest $\{ \rm{PosteriorMedian}(\rho_i) \}_{i=1}^D$.
    Note that there are three regularization hyperparameters in total.
  {\bf Right}: We depict the mean effective subspace dimension, defined to be the number of dimensions
    for which ${\rm{PosteriorMedian}(\rho_k) > \xi}$ where $\xi \in \{0.1, 0.5\}$ is an arbitrary cutoff.
    Means are averages across $30$ independent replications. }
  \label{fig:svm388relevance}
\end{figure}

In Fig.~\ref{fig:svm388relevance} we explore the relevant subspace identified by \algoname{}
during the course of optimization of the SVM problem discussed in Sec.~\ref{sec:sec_svm}.
We see that the three most important hyperparameters, namely the regularization hyperparameters,
are consistently found more or less immediately once the initial Sobol phase of Algorithm \ref{algo}
is over.
This explains the rapid early progress that \algoname{} makes in Fig.~\ref{fig:real_world}
during optimization. We note that the $4^{\rm th}$ most relevant dimension turns out to be a length
scale for a patient ID feature, which makes sense given the importance of this feature to the
regression problem.

\subsection{SVM ablation study}
\begin{figure}[!ht]
  \centering
  \includegraphics[width=0.99\linewidth]{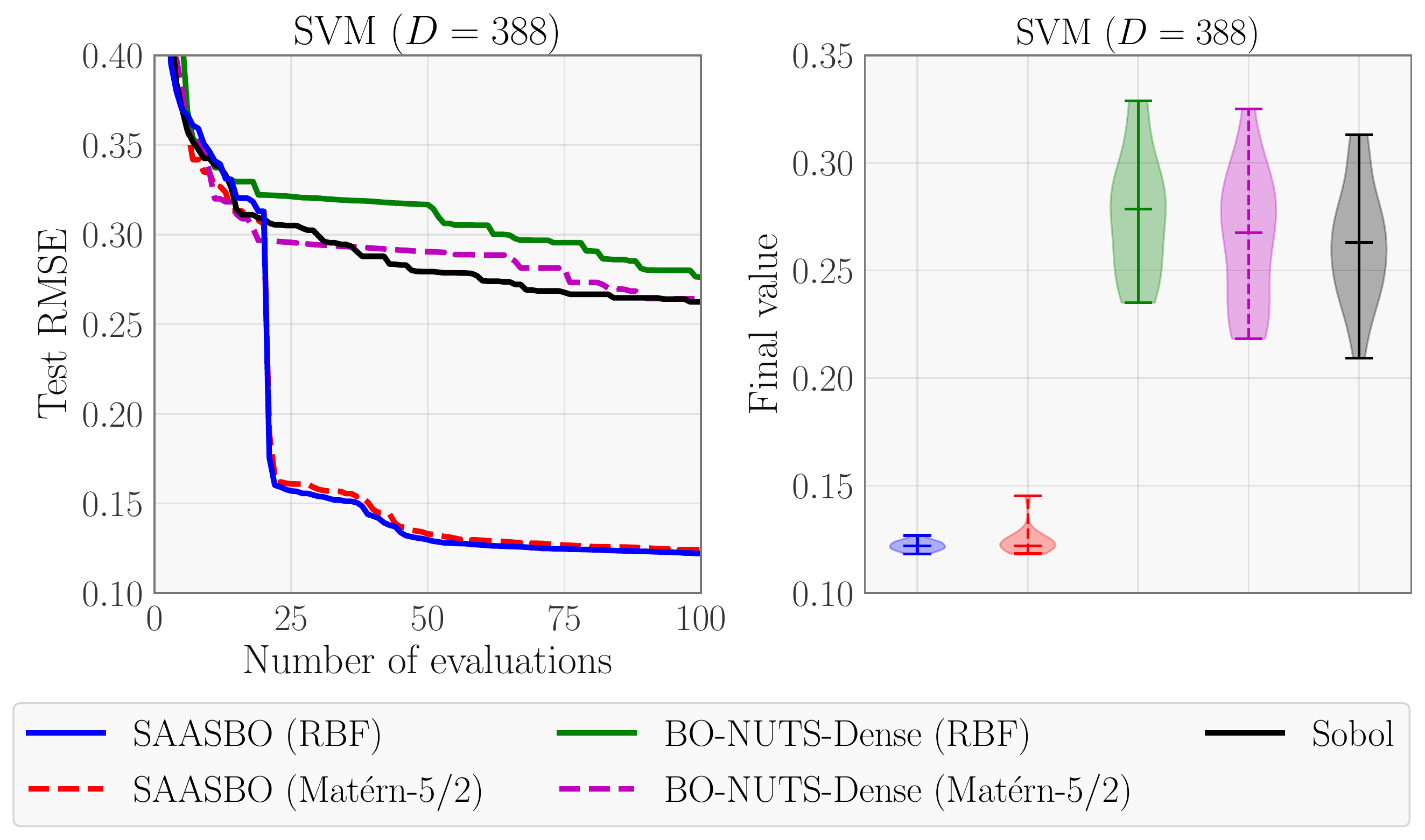}
    \caption{We compare the BO performance of the \priorname{} function prior to a non-sparse function prior
             on the SVM hyperparameter tuning problem ($D=388$). In addition we compare the RBF kernel
             to the Mat\'ern-$5/2$ kernel.
             We do $15$ independent replications for each method, except for \algoname-\texttt{RBF} and Sobol,
             for which we reproduce the same $30$ replications from the main text.
    {\bf Left:} For each method we depict the mean value of the best minimimum found at a given iteration.
    {\bf Right:} For each method we depict the distribution over the final approximate minimum $y_{\rm min}$ encoded as
    a violin plot, with horizontal bars corresponding to $5$\%, $50$\%, and $95$\% quantiles.
    }
    \label{fig:svm_ablation}
\end{figure}

In Fig.~\ref{fig:svm_ablation} we depict results from an ablation study of \algoname{} in the context of the SVM problem.
First, as a companion to Fig.~\ref{fig:model_cv} and Fig.~\ref{fig:model_cv_matern}, we compare the BO performance
of the \priorname{} function prior to a non-sparse function prior that places weak priors on the length scales.
As we would expect from Fig.~\ref{fig:model_cv} and Fig.~\ref{fig:model_cv_matern}, the resulting BO performance is
very poor for the non-sparse prior.
Second, we also compare the default RBF kernel to a Mat\'ern-$5/2$ kernel. We find that, at least on this problem,
both kernels lead to similar BO performance.

\subsection{Rotated Hartmann}
\label{sec:rotated}
In this experiment we study whether the axis-aligned assumption in \priorname{} leads to degraded performance
 on non-axis-aligned objective functions.
In particular, we consider the Hartmann function $f_{\text{hart}}$ for $d=6$ embedded in $D=100$ dimensions.
Given a projection dimensionality $d_p \geq d$, we generate a random linear projection $P_{d_p} \in \mathbb{R}^{d \times d_p}$ where $[P_{d_p}]_{ij} \sim \mathcal{N}(0, 1 / d_p)$.
The task is to optimize $\tilde{f}(\bx) = f_{\text{hart}}(P_{d_p} \bx_{1:d_p} - \bz))$ where $\bx \in [0, 1]^D$ and $\bz \in \mathbb{R}^d$.
For a given $P_{d_p}$, $\bz$ is a vector in $[0, 1]^{d}$ that satisfies $\tilde{f}([\bx^*; w]) = f_{\text{hart}}(\bx^*), \forall w \in [0, 1]^{D - d}$ where $\bx^*$ is the global optimum of the Hartmann function.
The translation $\bz$ guarantees that the global optimum value is attainable in the domain.
We consider $d_p \in \{ 6, 18, 30 \}$ and generate a random $P_{d_p}$ and $\bz$ for each embedded dimensionality;
these are then used for all replications.
EBO is excluded from this study, as it performed worse than Sobol in Fig.~\ref{fig:synthetic}.

\begin{figure*}[!t]
  \centering
  \includegraphics[width=0.98\linewidth]{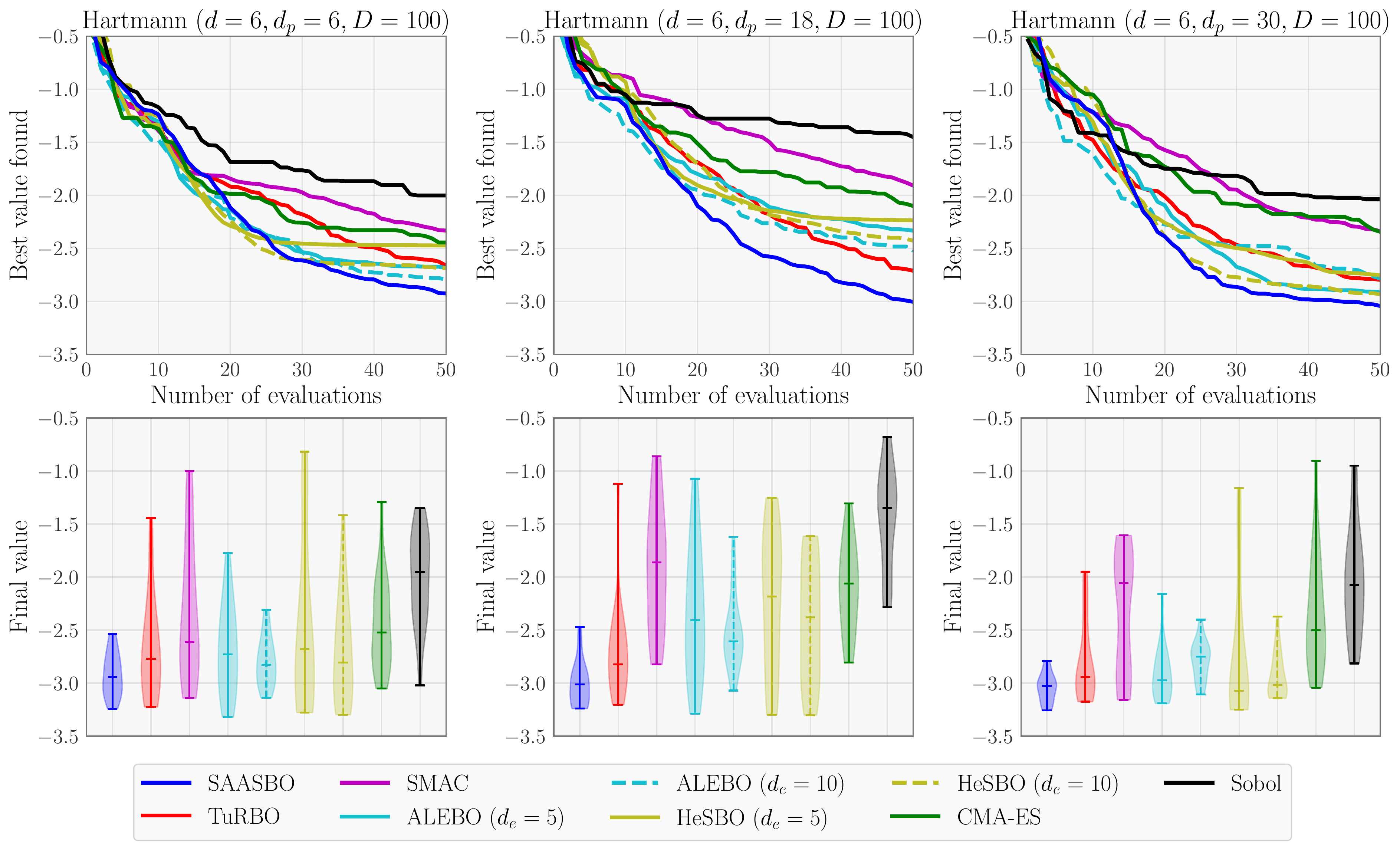}
    \caption{
      We consider a rotated version of the Hartmann function with $d=6$.
      We generate random linear projection matrices $P_{d_p}$ for different projection dimensionalities $d_p \in \{ 6, 18, 30 \}$ and optimize the resulting rotated function.
      \algoname{} outperforms the other methods and is able to quickly identify the most important parameters in the rotated coordinate system.
    }
    \label{fig:rotated_hartmann}
\end{figure*}

The results are shown in Fig.~\ref{fig:rotated_hartmann}.
We see that \algoname{} outperforms the other methods even though the function has been rotated, thus straining the axis-aligned assumption.
Despite the rotation, \algoname{} quickly identifies the most important parameters in the rotated space.
We also notice that the worst-case performance of \algoname{} is better than for the other methods across all projection dimensionalities considered.

\section{Additional Experimental Details}
\label{sec:expdetails}
Apart from the experiment in Sec.~\ref{sec:relevance} that is depicted
in Fig.~\ref{fig:branin3alphas} we use $\alpha=0.1$ in all experiments.
Apart from Fig.~\ref{fig:model_cv_matern} and Fig.~\ref{fig:svm_ablation}, we use an RBF kernel in all experiments.

\subsection{Model Fit Experiment}
In the model fit experiment in Sec.~\ref{sec:fit} we take data collected from two different runs of \algoname{} in $D=100$.
We use one run as training data and the second run as test data, each with $N=100$ datapoints.
To construct datasets in $D=30$ dimensions we include the $6$ relevant dimensions as well as $24$
randomly chosen redundant dimensions and drop all remaining dimensions.

\subsection{Inference and Hyperparameter Comparison Experiment}
For the experiment in Sec.~\ref{sec:relevance} that is depicted
in Fig.~\ref{fig:branin3alphas} we initialize \algoname{} with $m=10$
points from a Sobol sequence.

\subsection{Baselines}
We compare \algoname{} to ALEBO, CMA-ES, EBO, HeSBO, SMAC, Sobol, and TuRBO.
For ALEBO and HeSBO we use the implementations in BoTorch~\citep{balandat2020botorch} with the same settings that were
used by~\citep{letham2020re}.
We consider embeddings of dimensionality $d_e=5$ and $d_e=10$ on the synthetic problems, which is similar to the $d_e = d$ and $d_e = 2 d$ heuristics that were considered in~\citep{nayebi2019framework} as well as~\citep{letham2020re}.
As the true active dimensionality $d$ of $\fobj$ is unknown, we do not allow any method to explicitly use this additional information.
For the three real-world experiments, $d_e=5$ does not work well on any problem so we instead
report results for $d_e=10$ and $d_e=20$.

For CMA-ES we use the \texttt{pycma}\footnote{\url{https://github.com/CMA-ES/pycma}} implementation.
CMA-ES is initialized using a random point in the domain and uses the default initial step-size of $0.25$.
Recall that the domain is normalized to $[0, 1]^D$ for all problems.
We run EBO using the reference implementation by the authors\footnote{\url{https://github.com/zi-w/Ensemble-Bayesian-Optimization}} with the default settings.
EBO requires knowing the value of the function at the global optimum.
Similarly to~\citep{letham2020re} we provide this value to EBO for all problems, but note that EBO still performs poorly on all problems apart from Branin and SVM.

Our comparison to SMAC uses \texttt{SMAC4HPO}, which is implemented in \texttt{SMAC3}\footnote{\url{https://github.com/automl/SMAC3}}.
On all problems we run SMAC in deterministic mode, as all problems considered in this paper are noise-free.
For Sobol we use the \texttt{SobolEngine} implementation in PyTorch.
Finally, we compare to TuRBO with a single trust region due to the limited evaluation budget; we use the implementation provided by the authors\footnote{\url{https://github.com/uber-research/TuRBO}}.

\subsection{Synthetic problems}
We consider three standard synthetic functions from the optimization literature.
Branin is a $2$-dimensional function that we embed in a $100$-dimensional space.
We consider the standard domain $[-5, 10] \times [0, 15]$ before normalizing the
domain to $[0, 1]^{100}$.
For Hartmann, we consider the $d=6$ version on the domain $[0, 1]^6$ before embedding it in a $100$-dimensional space.
For Rosenbrock, we use $d=3$ and the domain $[-2, 2]^3$, which we then embed and normalize so that the full domain is $[0, 1]^{100}$.
Rosenbrock is a function that is challenging to model, as there are large function values at the boundary of the domain.
For this reason all methods minimize ${\log(1 + \fobj(x))}$.
All methods except for CMA-ES are initialized with $m=10$ initial points for Branin and Rosenbrock and
$m=20$ initial points for Hartmann.

\subsection{Rover}
We consider the rover trajectory optimization problem that was also considered in~\citet{wang2018batched}.
The goal is to optimize the trajectory of a rover where this trajectory is determined by fitting a B-spline to $30$ waypoints in the $2$D plane.
While the original problem had a pre-determined origin and destination, the resulting B-spline was not constrained to start and end at these positions.
To make the problem easier, we force the B-spline to start and end at these pre-determined positions.
Additionally, we use $50$ waypoints points, which results in a $100$-dimensional optimization problem.
The reward function for the trajectory is computed in the same way as in~\citet{wang2018batched},
namely we integrate over the trajectory penalizing collisions with potential objects.
On this problem we initialize all methods except for CMA-ES with $m=20$ initial points.

\subsection{SVM}
We randomly choose $5000$ training and $5000$ test points
from the $385$-dimensional ``CT slice''\footnote{\url{https://archive.ics.uci.edu/ml/datasets/Relative+location+of+CT+slices+on+axial+axis}} UCI dataset~\citep{dua2019uci}.
We normalize the inputs and scalar output so that e.g.~the test RMSE of a trivial
zero prediction is given by $1.0$.
Our domain $\DD$ then consists of $385$ kernel (log) length scales and $3$ regularization hyperparameters
for a kernel support vector machine fit with \texttt{Scikit-learn} \citep{pedregosa2011scikit}. The log length scales
are restricted to the interval $[-2, 2]$. The 3 regularization hyperparameters,
which are likewise represented in log space,
are denoted \texttt{epsilon}, \texttt{C}, and \texttt{gamma} in the \texttt{SVR} class constructor. We restrict \texttt{epsilon} to $[0.01, 1.0]$, \texttt{gamma} to $[0.1, 3.0]$,
and \texttt{C} to $[0.01, 5.0]$. Aftering fitting the SVM regressor to the training
data we compute the test RMSE (root mean squared error). This test RMSE is the quantity
we seek to minimize. We use the default settings of \texttt{SVR}, which among other things
means the kernel used is a RBF kernel.
On this problem we initialize all methods except for CMA-ES with $m=20$ initial points.

\subsection{MOPTA Vehicle Design}
We consider the vehicle design problem MOPTA08 which is a challenging $124$-dimensional real-world high-dimensional BO problem~\citep{jones2008large}.
The goal in this problem is to minimize the mass of a vehicle subject to $68$ performance constraints.
The $D=124$ design variables describe materials, gauges, and vehicle shape.
While this problem is originally formulated as a constrained optimization problem, we make it unconstrained by converting the constraints into a soft constraint.
In particular, we consider minimizing ${\fobj(x) + 10 \sum_{i=1}^{68} \max(0, c_i(x))}$ where the $68$ constraints are of the form $c_i(x) \leq 0$.
This penalty is chosen to be small enough to have most of the signal come from $\fobj$ while at the same time discouraging large constraint violations.
While it is worth emphasizing that there are constrained optimization methods that can explicitly handle the constraint, this problem shows that \algoname{} can quickly exploit structure in $\fobj$ even though there is no obvious low-dimensional structure.

For \algoname{} we use the NUTS settings described in Sec.~\ref{sec:suppnuts} for $t\le150$.
To lower the runtime after iteration $t>150$ we collect $384=192+192$ NUTS samples and retain every $24^{\rm th}$ of the final $192$ samples, resulting in a total of $L=8$ retained samples.
We note while this may hurt the accuracy of the inferred GP model, \algoname{} still performs
very well on this problem and outperforms other methods by a large margin.
As we consider a larger evaluation budget on this problem we initialize all methods except for CMA-ES with $m=50$ initial points.

\end{document}